\documentclass[review]{elsarticle}
\usepackage{amsfonts} 
\usepackage{hyperref}
\usepackage{color,xcolor}
\usepackage{verbatim}
\usepackage{multirow}
\usepackage{times}
\usepackage{epsfig}
\usepackage{graphicx}
\usepackage{amsmath}
\usepackage{amssymb}
\usepackage{makecell}
\usepackage{bm, xspace}

\journal{Journal of NEUROCOMPUTING Templates}

\makeatletter
\DeclareRobustCommand\onedot{\futurelet\@let@token\@onedot}
\def\@onedot{\ifx\@let@token. \else.\null\fi\xspace}

\def\eg{\emph{e.g}\onedot}

\def\etc{\emph{etc}\onedot} 
\def\wrt{w.r.t\onedot} 
\def\etal{\emph{et al}\onedot} 
\makeatother

\begin{document}
\graphicspath{{images/}}

\begin{frontmatter}

\title{A Holistic Representation Guided Attention Network for Scene Text Recognition}

\author[NWPU,ASGO]{Lu Yang}
\ead{lu.yang@mail.nwpu.edu.cn}

\author[NWPU,ASGO]{Fan Dang}
\ead{dangfan@mail.nwpu.edu.cn}

\author[NWPU,ASGO]{Peng Wang\corref{corresponding_author}}
\cortext[corresponding_author]{Corresponding author}
\ead{peng.wang@nwpu.edu.cn}

\author[Adelaide]{Hui Li}
\ead{huili03855@gmail.com}
\author[MinSheng]{Zhen Li}
\ead{lizhen@mskj.com}
\author[NWPU,ASGO]{Yanning Zhang}
\ead{ynzhang@nwpu.edu.cn}

\address[NWPU]{School of Computer Science, Northwestern Polytechnical University, Xi'an, China}
\address[ASGO]{National Engineering Laboratory for Integrated Aero-Space-Ground-Ocean Big Data Application Technology, China}
\address[Adelaide]{School of Computer Science, The University of Adelaide, Australia}
\address[MinSheng]{MinSheng FinTech Corp. Ltd., China}

\begin{abstract}
Reading irregular scene text of arbitrary shape in natural images is still a challenging problem, despite the progress made recently.
Many existing approaches incorporate sophisticated network structures to handle various shapes,
use extra annotations for stronger supervision,
or employ hard-to-train recurrent neural networks for sequence modeling.
In this work, we propose a simple yet strong approach for scene text recognition.
With no need to convert 
input images to sequence representations, we directly connect two-dimensional CNN features to an attention-based sequence decoder which guided by holistic representation. 
The holistic representation can guide the attention-based decoder focus on more accurate area. 
As no recurrent module is adopted, our model can be trained in parallel.  
It achieves $1.5\times$ to $9.4\times$ acceleration to backward pass and $1.3\times$ to $7.9\times$ acceleration to forward pass, compared with the RNN counterparts.
The proposed model is trained with only word-level annotations.
With this simple design, our method 
achieves state-of-the-art or competitive recognition performance on the evaluated regular and irregular scene text benchmark datasets. 
\end{abstract}


\end{frontmatter}

\section{Introduction}
Text in natural scene images contains rich semantic information that is crucial for visual understanding and reasoning in many cases.
Text reading has been integrated in a variety of vision tasks, such as fine-grained image classification~\cite{karaoglu2017words,karaoglu2017text,bai2018integrating},
image retrieval~\cite{karaoglu2017words,gomez2018single} and visual question answering~\cite{gurari2018vizwiz,textvqa}.

\begin{figure*}[t!]
	\begin{center}
		\includegraphics[width=1.0\textwidth]{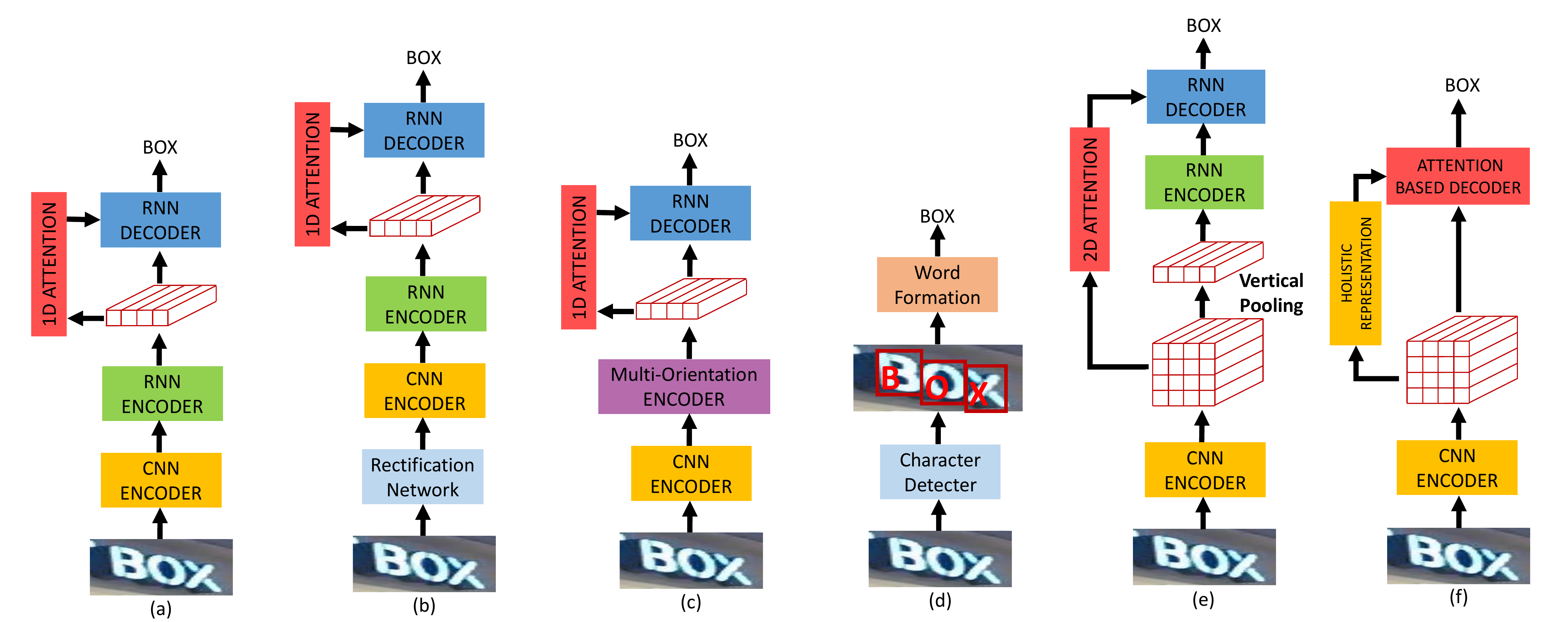}
	\end{center}
	\vspace{-5mm}
	\caption{Typical architectures and our model for scene text recognition. (a) is the basic $1$D attention based encoder-decoder framework for regular text recognizer~\cite{Lee_2016_CVPR}. (b)-(f) are all for irregular text recognition. (b) Shape rectification based~\cite{shiPAMI2018}; (c) Multi-direction encoding based~\cite{Cheng2018AON}; (d) Character detection based~\cite{Liao2019}; (e) $2$D attention based~\cite{Li_AAAI2019}; (f) our model (end-to-end trainable convolutional attention based, without RNNs being used.)
	}
	\label{fig:intro}
\end{figure*}

Recognizing regular text in almost straight lines can be considered as a sequence-to-sequence problem and solved by an attentional Recurrent Neural Network (RNN) framework as shown in Figure~\ref{fig:intro}(a).
In comparison to regular text recognition, it is much more challenging to 
recognize 
irregular text of arbitrary shape for a machine. 
Existing approaches for irregular text recognition can be roughly categorized into four types, namely, 
shape rectification, multi-direction encoding, character detection and $2$D attention based approaches, as shown in Figure~\ref{fig:intro}(b), (c), (d), (e) respectively. 
The shape rectification based methods~\cite{shiPAMI2018} first approximately rectify irregular text into regular one, and then apply regular text recognizers. Nevertheless, severely distorted or curved shapes are difficult to be rectified. 
Cheng~\etal~\cite{Cheng2018AON} propose a sophisticated four-directional encoding method to recognize arbitrarily-oriented text, which, however, introduces redundant representations. 
Character detection based methods~\cite{Liao2019} firstly detect and recognize individual characters and then connect them using a separate post-processing method, which inherently requires character-level annotations and cannot be trained end-to-end.
$2$D attention based approaches 
learn to 
focus on
individual character features in $2$D spaces during decoding, which can be trained either with word-level~\cite{Li_AAAI2019} or character-level annotations~\cite{ijcai2017}.

Note that a large number of irregular text recognizers (\eg, \cite{shiPAMI2018,Cheng2017,Cheng2018AON,Li_AAAI2019,shiCVPR2016}) still need to convert input images into intermediate sequence representations, and use RNNs to encode and decode them.
There are two limitations for this type of approaches.
First, given that irregular text actually being distributed in two dimensional spaces, to some extent, it is inappropriate and difficult to convert them into one dimensional sequence representations.
As shown in \cite{Liao2019}, solving the irregular text recognition problem from two dimensional perspective may yield more robust performance.
Second, RNNs are inherently difficult to be parallelized and 
typically 
hard to train due to the problem of gradient vanishing/exploding. 
In the field of regular text recognition, some attempts have been made to replace RNNs with non-recurrent architectures, including convolution based~\cite{gao2019reading} and attention based sequence modeling~\cite{sheng2018nrtr} methods.
However, both methods are still based on sequence-to-sequence structures,
which is 
not well capable of 
handling irregular text of arbitrary shape.  

To this end, we propose a simple yet 
robust architecture for irregular text recognition, as shown in Figure~\ref{fig:intro}(f).
Our approach  directly connects a CNN-based $2$D image encoder to an attention-based $1$D sequence decoder, preventing from using intermediate sequence representations.  
Inspired by the Transformer~\cite{vaswani2017attention} in NLP, we adopt an attention-based decoder that does not rely on recurrent connections and so can be trained in parallel and converges quickly.

Note that the Transformer is proposed for machine translation, taking $1$D sequences as inputs. But the inputs of our proposed irregular text recognizer are $2$D images, 
which makes these two models different from each other.   
The self-attention mechanism, which plays a key role in the Transformer to model long-range dependencies in both input and output sequences, is relatively less important in our model for text recognition. 

Firstly, instead of using self-attention, we use a CNN to encode input scene text images.
Accordingly, we need to use $2$D attention in the decoder.
Secondly, 
the employment of self-attention in the decoder offers no significant performance gain. 
This is not surprising: the dependency between characters of a single word is typically weaker than that between words of a sentence or paragraph. 

Our main contributions are three-fold:

\noindent{\bf 1)} The proposed model is simple by design. It only consists of a 
CNN model for image encoding and a tailored attention-based sequence decoder.  
Unlike sequence-to-sequence text recognizers, we do not convert input images to sequence representations, which itself is challenging for text of complex shape.
Instead, we convert the input image to a $2$D feature map and a $1$D holistic representation by a CNN model, and then connect them directly to the sequence decoder.
Furthermore, the training of the proposed model only requires word-level annotations, which enables it to be trained with real data that usually does not come with character-level annotations. 

\noindent{\bf 2)} Our proposed method is an end-to-end trainable non-recurrent network for both regular and irregular text recognition. Without using any RNN module, this model can be trained in parallel.
Compared with state-of-the-art RNN-based irregular text recognizers~\cite{shiPAMI2018,Li_AAAI2019}, our model is $1.5\times$ to $9.4\times$ faster in backward pass and $1.3\times$ to $7.9\times$ faster in forward pass. This acceleration leads to a rapid experimental turnaround and makes our model scalable to larger datasets.

\noindent{\bf 3)} We encodes the rich context information of the entire image as a holistic representation, and the algorithm focuses better on the characters to be decoded with the context information provided by the holistic representation. As Figure~\ref{fig:holisticCase} shows, the algorithm focuses on more accurate area with the help of holistic representation. 

\paragraph{\bf Notation}
Matrices and column vectors are denoted by bold upper and lower case letters respectively.
$\mathbb{R}^m$ and $\mathbb{R}^{m \times n}$ indicate real-valued $m$ dimensional vectors and $m \times n$ matrices respectively.
$\langle \mathbf{a}, \mathbf{b} \rangle \in \mathbb{R}$ means the inner-product of $\mathbf{a} \in \mathbb{R}^m $ and $\mathbf{b} \in \mathbb{R}^m $.
$[ \mathbf{a}, \mathbf{b} ] \in \mathbb{R}^{m \times 2}$ 
and 
$[ \mathbf{a}; \mathbf{b} ] \in \mathbb{R}^{2m}$
represent the horizontal and vertical stacks of $\mathbf{a}$ and $\mathbf{b}$ respectively.

\section{Related Work}

\paragraph{\bf{Irregular Scene Text Recognition}} \,\,\,
Early work for scene text recognition adopts a bottom-up fashion~\cite{Wangkai2011, SVTP}, which detects individual characters firstly~\cite{tang2018scene} and integrates them into a word by means of dynamic programming, or a top-down manner~\cite{Max2016IJCV}, which treats the word patch as a whole and recognizes it
as 
multi-class image classification. Considering that scene text generally appears in the form of a character sequence, recent work models it as a sequence recognition problem. RNNs are generally 
used
for sequential feature learning. 
Connectionist Temporal Classification (CTC) and sequence-to-sequence learning models are two prevalent methods that are widely used for scene text recognition~\cite{ShiBY15, Lee_2016_CVPR, Cheng2017, Hui2017ICCV, cheng_EditDistance}. 
Besides, some multilingual scene text recognition datasets and multilingual scene text recognition work has also been studied~\cite{tian2016multilingual, Shi2017ICDAR2017}.



Methods for irregular text recognition are mostly driven by the above frameworks but involve some improvements to deal with the distortions or curvatures of irregular text. 
For instance, Su and Lu~\cite{Su2017Accurate} proposed a new ensembling technique that combines outputs from two RNNs for better recognition results, and use HoG instead of deep network features for recognition.
Shi~\etal~\cite{shiCVPR2016, shiPAMI2018} proposed to rectify irregular text images into regular ones by Spatial Transformer Network (STW)~\cite{jaderberg2015spatial},
and then recognized them using a $1$D attentional sequence-to-sequence model. Zhan and Lu~\cite{ESIR} proposed to iteratively remove perspective distortion and text line curvature by an innovative rectification network so as to result in a fronto-parallel view of text for recognition.
Rather than rectifying the entire word image, Liu~\etal~\cite{Liu2018CharNetAC} proposed to detect and rectify individual characters in the word by STW. 
Cheng~\cite{Cheng2018AON} captured the deep features of irregular text image along four directions by RNNs, which are then combined by $1$D attention based decoder to generate character sequence. A filter gate was designed to fuse those redundant features and remove irrelevant ones. 
Liao~\etal~\cite{Liao2019} argued that it is inappropriate to represent irregular text image with a $1$D sequence, and proposed a Character Attention Fully Convolutional Network to detect each character accurately in two-dimensional perspective. Word formation is then realized with a separate segmentation based method. This model cannot be trained end-to-end.
Some methods attempt to extend $1$D attention mechanism into $2$D spaces. 
Character-level annotations are 
often needed 
to supervise the training of $2$D attention network. For example, the Focusing Attention Network (FAN) proposed by Cheng~\etal~\cite{Cheng2017} introduced a focus network to tackle the attention drift between the local character feature and target. Yang~\etal~\cite{ijcai2017} introduced an auxiliary Fully Convolutional Network for dense character detection. An alignment loss was used to supervise the training of attention model during word decoding. 
Li~\etal~\cite{Li_AAAI2019} modified the attention model and proposed a tailored $2$D attention based framework for exact local feature extraction. 
Nevertheless, $2$-layer RNNs are adopted respectively in both encoder and decoder which precludes computation parallelization and suffers from heavy computational burden.

\paragraph{\bf{Non-recurrent Sequence Modeling}} \,\,\,
Some work has been proposed in recent years to remove the recurrent structure in the  sequence-to-sequence learning framework, so as to enable fully parallel computation and accelerate the processing speed. Gehring~\etal~\cite{gehring2017convolutional} proposed an architecture for machine translation with entirely convolutional layers. Compositional structures in the sequence can be discovered based on the hierarchical representations. However, this model still has difficulty to learn dependencies between distant positions. 
Vaswani~\etal~\cite{vaswani2017attention} proposed a ``Transformer'' for machine translation, which is based solely on attention mechanisms. The fundamental self-attention module can draw dependencies between different positions in a sequence through position-pair computation rather than position-chain computed by RNNs, which leads to more computation parallelization and less model complexity. 
Inspired 
by this model, Dong~\etal~\cite{Dong2018} introduced Transformer to speech recognition and Yu~\etal~\cite{QANet} combined local convolution with global self-attention for reading comprehension task. Most recently, Dehghani~\etal~\cite{dehghani2018universal} generalized the Transformer and proposed the ``Universal Transformer'' to deal with string copying or logical inference with string's length exceeding those observed at training time. There are also some efforts for scene text reading without using recurrent networks. Gao~\etal~\cite{gao2019reading} presented an end-to-end attention convolutional network for scene text recognition, with a CTC layer followed to generate the final label. Wu~\cite{SCAN} presented a sliding convolutional attention network for scene text recognition, based on the convolutional sequence-to-sequence learning framework~\cite{gehring2017convolutional}. Sheng~\etal~\cite{sheng2018nrtr} proposed a non-recurrent sequence-to-sequence model for scene text recognition based on Transformer~\cite{vaswani2017attention}, with self-attention module working as the basic block in both encoder and decoder to learn character dependencies.
All these sequence-to-sequence frameworks are mainly for regular text recognition and 
are 
not easy to be extended 
to handle 
irregular text because of their inherent model design. In contrast, in this work, we propose an simple yet effective $2$D image to $1$D sequence model based on convolution and attention modules. It maps text images into character sequences directly and can address both regular and irregular scene text recognition. 

\section{Model Architecture}

As shown in Figure~\ref{fig:framework}, the proposed model is based on an encoder-decoder structure, which is popular for many cross-modality transformation tasks. 
Previous sequence-to-sequence based text recognizers represent input images with 
1D sequences, which, however, encounter difficulties when dealing with irregular text scattering in $2$D spaces.
Alternatively, we employ a CNN encoder to extract both $2$D feature map (two dimensional representations) and holistic representation (one dimensional representation) of text images.
The resulting image representations are then fed into 
an attention-based decoder with a stack of masked self-attention, $2$D attention and point-wise feed-forward layers.

During testing, the decoder takes as input at each step the concatenation of the holistic representation and the embedding of the previously generated character which is added with the encoding of the current position,
adaptively focuses on the related image regions via $2$D attention,
and predict the character at the current position.
During training, the computation of the decoder can be 
easily parallelized with given ground-truth labels. 
In the following, we will introduce each component of our proposed model in detail. 

\subsection{Encoder}
\begin{figure*}[tbp]
	\begin{center}
		\includegraphics[width=0.9\textwidth]{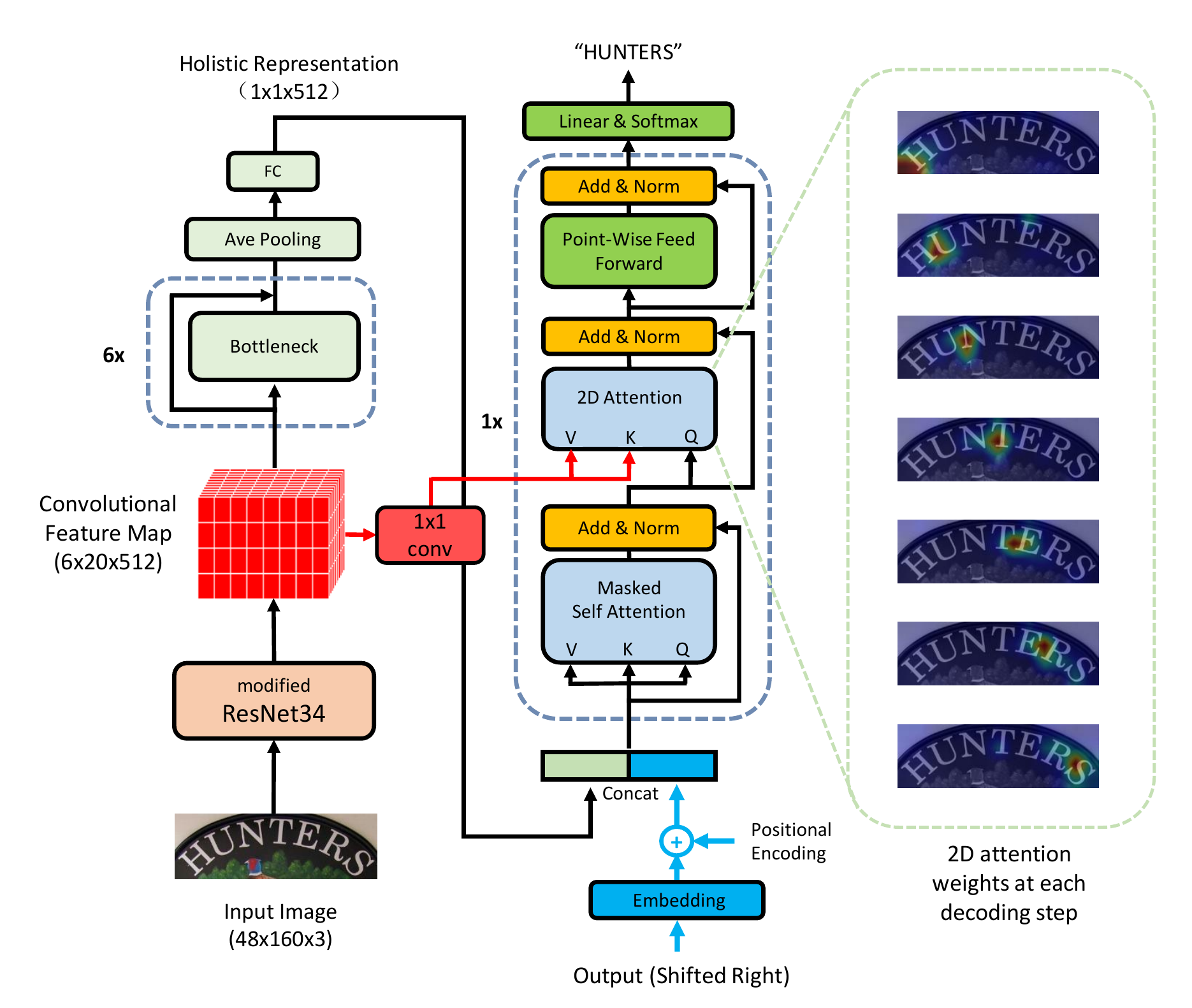}
	\end{center}
	\vspace{-1em}
	\caption{The overall structure of our proposed model. It consists of two parts: a ResNet34-based image encoder (left) and an attention-based sequence decoder (right). The $2$D feature maps generated by modified ResNet34 are connected to the $2$D attention module in the decoder by a $1 \times 1$ convolution layer, and we stack 6 bottleneck modules to extract the $1$D holistic representation which can help the $2$D attention more accurate. The bottleneck module is the same as in ResNet~\cite{ResidualNetwork}.
	In contrast to other irregular text recognizers ~\cite{Luo2019MORAN,shiPAMI2018}, there is no recurrent networks to model the representation.
	As a non-recurrent network, our model can be trained in parallel. Furthermore, training our model only needs word-level annotations.
	}
	\label{fig:framework}
\end{figure*}

\vspace{-0.5em}
We adopt as our CNN encoder the ResNet34~\cite{ResidualNetwork}
based architecture, which consists of a modified ResNet34 and a holistic representation extractor as shown in Figure~\ref{fig:framework}. The detail information of the modified ResNet34 is shown in Table~\ref{tab:modified_resnet}.
The final average pooling layer of the ResNet34 is removed and then followed two branches. One branch is a $1 \times 1$ convolution layer to transform the dimension of the $2$D feature map from $512$ to $1024$ and feed into $2$D attention, the other branch is a holistic representation extractor which consists of $B$ bottlenecks, average pooling and a fully connected layer. The resulting holistic representation constitute the input of the decoder. The ablation study shows that $B=6$ is enough in our case (see Section~\ref{sec:ablation} for details).
All the input images are uniformly resized into $48 \times 160 \times 3$, resulting in feature maps of size $6 \times 20 \times 512$.
We also evaluate other CNN backbones such as ResNet18, ResNet50 and ResNet152 for image encoding, which do not offer significant performance improvements, as referred to the ablation experiments. 
Note that it may be more reasonable to rescale images without destroying their original aspect ratios~\cite{Li_AAAI2019}, which we leave for future work.

\begin{table}[]
\begin{center}
\scalebox{0.8}{
\begin{tabular}{c|c}
\hline
Layer Name  & Configuration \\ \hline \hline
Conv        & $3\times3$, $64$        \\ \hline
Batch Norm  &  $-$  \\ \hline
Relu        &  $-$  \\ \hline
Maxpool1    &  k: $2\times2$, s: $2\times2$   \\ \hline
Layer1      &  Basicblock$\times3$, $64$, s: $1\times1$  \\ \hline
Maxpool2    &  k: $2\times2$, s: $2\times2$   \\ \hline
Layer2      &  Basicblock$\times4$, $128$, s: $1\times1$ \\ \hline
Maxpool3    &  k: $2\times2$, s: $2\times2$   \\ \hline
Layer3      &  Basicblock$\times6$, $256$, s: $1\times1$ \\ \hline
Layer4      &  Basicblock$\times3$, $512$, s: $1\times1$ \\ \hline
\end{tabular}
}
\end{center}
\vspace{-1mm}
\caption{The configuration of modified ResNet34. ``Conv'' stands for Convolutional layers, with kernel size
and output channels presented. The stride and padding for convolutional layers are all set to ``1''. ``Layer'' stands for stacked basic blocks, with the number of blocks, output channels and stride presented. ``k'' means kernel size, and ``s'' represents stride. }
\vspace{-1em}
\label{tab:modified_resnet}
\end{table}

\subsection{Decoder}
Inspired by \cite{vaswani2017attention}, the designed attention-based sequence decoder is composed of three layers: 
1) a masked self-attention mechanism for modeling dependencies between different characters within output words;
2) a $2$D attention module linking encoder and decoder;
and 
3) a point-wise feed-forward layer applied to each decoding position separately.
A residual connection with an addition operation is employed for each of the above three layers, followed by layer normalization. 
The above three components form a block and can be stacked $N$ times without sharing parameters.
There are $N=6$ blocks in the Transformer~\cite{vaswani2017attention}, but we found that using only one block already achieves saturated performance in our case (see Section~\ref{sec:ablation}). 
In the following, we describe the decoder components in detail.

\paragraph{\bf Output Embedding and Positional Encoding} \,\,
During testing, the previously generated character will be embedded to a $d/2$-dimensional vector at each decoding step,
which is further added with the encoding of the current position as follows:
\begin{equation}
\label{eq:position}
\mathrm{PE}(p,i) = \left\{ \begin{array}{ll}
\mathrm{sin}(p/10000^{i/(d/2)})   & \mbox{if} \,\, i \,\, \mbox{is even}   \\
\mathrm{cos}(p/10000^{(i-1)/(d/2)}) & \mbox{if} \,\, i \,\, \mbox{is odd}
\end{array}\right. 
\end{equation}
where $p$ is the position and $i \in \{1, \dots, (d/2)\}$ is the dimension. Then they are concatenated to a holistic representation.
While at training time, the ground-truth characters are shifted right and embedded simultaneously, which enables parallel training.

\paragraph{\bf Multi-Head Dot-Product Attention} \,\,
Both masked self-attention and two-dimensional attention in our decoder are based on the multi-head dot-product attention formulation~\cite{vaswani2017attention}. Here, we briefly review this formulation. 
The scaled dot-product attention takes as inputs a query $\mathbf{q} \in \mathbb{R}^d$ and a set of key-value pairs of $d$-dimensional vectors $\{(\mathbf{k}_i, \mathbf{v}_i)\}_{i=1,2,\dots,M}$ ($M$ is the number of key-value pairs),
and computes as output a weighted sum of the values, where the weight for each value is computed by a scaled dot-product of the query and the corresponding key.
The formulation of scaled dot-product attention can be expressed as follows:\\
\begin{align}
\label{eq:dotprod}
&\mathrm{Atten(\mathbf{q},\mathbf{K},\mathbf{V})} = \sum_{i=1}^{M} \alpha_i \mathbf{v}_i \in \mathbb{R}^d \\
&\mbox{where} \,\,\, \boldsymbol{\alpha} = \mathrm{softmax} \big( \frac{\langle \mathbf{q}, \mathbf{k}_1 \rangle}{\sqrt{d}}, \frac{\langle \mathbf{q}, \mathbf{k}_2 \rangle}{\sqrt{d}}, 
  \dots, \frac{\langle \mathbf{q}, \mathbf{k}_M \rangle}{\sqrt{d}} \big) \notag
\end{align}

is the attention weights, $\mathbf{K} = [\mathbf{k}_1, \mathbf{k}_2, \dots, \mathbf{k}_M]$ and $\mathbf{V} = [\mathbf{v}_1, \mathbf{v}_2, \dots, \mathbf{v}_M]$.   
If there is a set of queries $\mathbf{Q} = [\mathbf{q}_1, \mathbf{q}_2, \dots, \mathbf{q}_{M'}]$ ($M'$ is the number of queries), 
then we have:
\begin{align}
\label{eq:dotprod1}
&\mathrm{Atten(\mathbf{Q},\mathbf{K},\mathbf{V})} = [\mathbf{a}_1, \mathbf{a}_2, \dots, \mathbf{a}_{M'}] \in \mathbb{R}^{d \times {M'}}\\
&{\mbox{where}} \,\,\, \mathbf{a}_i =  \mathrm{Atten(\mathbf{q}_i,\mathbf{K},\mathbf{V})}. \notag
\end{align}
The above scaled dot-product attention can be applied multiple times (multi-head) with different linear projections to $\mathbf{Q}$, $\mathbf{K}$ and $\mathbf{V}$, followed by a concatenation
and projection operation:
\begin{align}
\label{eq:dotprod2}
&\mathrm{MHAtten(\mathbf{Q},\mathbf{K},\mathbf{V})} = \mathbf{W}^o[\mathbf{A}_1; \dots; \mathbf{A}_H] \in \mathbb{R}^{d \times {M'}} \\
&{\mbox{where}} \,\,\, \mathbf{A}_i = \mathrm{Atten(\mathbf{W}_i^q \mathbf{Q}, \mathbf{W}_i^k \mathbf{K}, \mathbf{W}_i^v \mathbf{V})}. \notag
\end{align}
The parameters are $\mathbf{W}_i^q \in \mathbb{R}^{\frac{d}{H} \times d}$, $\mathbf{W}_i^k \in \mathbb{R}^{\frac{d}{H} \times d}$, $\mathbf{W}_i^v \in \mathbb{R}^{\frac{d}{H} \times d}$ 
and $\mathbf{W}^o \in \mathbb{R}^{d \times d}$.  
We set the number of attention heads $H$ to $16$ for our proposed model (see Section~\ref{sec:ablation} for the ablation study on the selection of $H$).

\paragraph{\bf Masked Self-Attention} \,\,
This attention layer is used to model the dependencies between different decoding positions,
where the queries, keys and values are the same. 
In this case, $M = M' = \mbox{the length of decoded sequence}$.
A mask is applied to prevent each position from attending to positions after that position.

\paragraph{\bf Two-Dimensional Attention} \,\,
In this layer, the queries come from the masked self-attention layer, and the keys and values are the $2$D output features of the CNN encoder.
In this case, $M = 6 \times 20$ and $M'$ is the length of decoded sequence.
It is the main connection between the encoder and decoder, which allows each decoding position attend to the $2$D positions of the encoder outputs.

\paragraph{\bf Point-wise Feed-Forward Layer} \,\,
A simple feed-forward network is applied at each position of the outputs of two-dimensional attention layer, 
which contains two linear transformations of dimension $d'$ and a ReLU non-linearity in between.  
The parameters of this layer are shared across all positions. 

\paragraph{\bf Prediction and Loss Function} \,\,
A linear transformation followed by a softmax function is used to transform the decoder output into prediction probabilities over character classes. Here we use $94$ character classes, including digits, case-sensitive letters and $32$ punctuation characters.
The parameters are also shared over all decoding positions.
The standard cross-entropy function is adopted to compute the loss of the predicted probabilities \wrt\  the ground-truth, at each decoding position.


\subsection{Bidirectional Decoder}
\begin{figure}[]
\small
	\begin{center}
		\includegraphics[width=8.0cm, height=2.5cm]{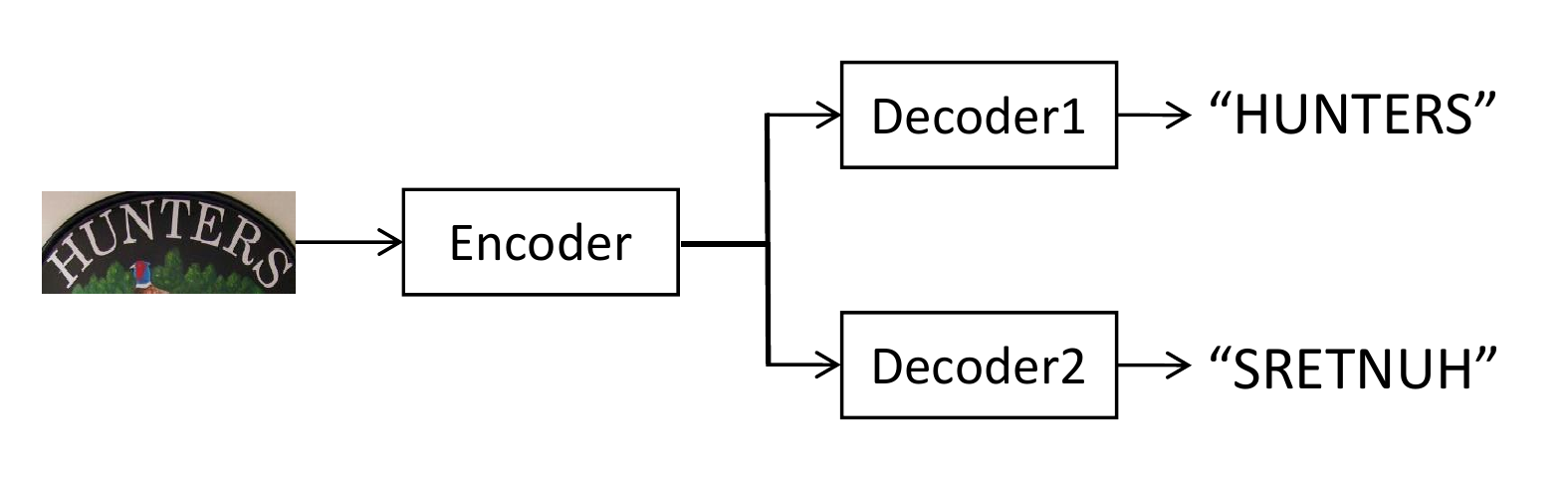}
	\end{center}
    \vspace{-1mm}	
	\caption{Bidirectional decoder in our model. The architecture of decoder1 and decoder2 is the same, and the parameters are not shared. The difference between them is the order of label sequences, one is normal and the other is reversed.}
	\vspace{-0.5em}
	\label{fig:BidirectionalDecoder}
\end{figure}

ASTER~\cite{shiPAMI2018} propose a bidirectional decoder, which consists of two decoders with opposite directions. As illustrated in Figure~\ref{fig:BidirectionalDecoder}, one decoder is trained to predict the characters from left to right and the other right to left. If a sample's first character is difficult to recognize, it may be easier by the help of right to left label. Because in reverse order, the first character becomes the last, then other characters will play a certain role in the recognition of the last character, so as to improve the recognition accuracy. In the test phase, we pick the one with the highest recognition score between the two decoders.

\section{Experiments}
The datasets for scene text recognition have been extensive studied in~\cite{baek2019what}, and we evaluate the performance of our method on a number of scene text recognition datasets consistent with ~\cite{baek2019what}. Ablation study is also conducted to investigate the impact of different model hyper-parameters.

\subsection{Datasets}
\begin{sloppypar}Our model is solely trained on synthetic datasets without using any real-world images. The same trained model, without further fine-tuning, is then evaluated on the following standard datasets: 
IIIT 5K-Words (IIIT5K)~\cite{MishraBMVC12}, 
Street View Text (SVT)~\cite{Wangkai2011}, 
ICDAR2013 (IC13)~\cite{icdar2013}, 
ICDAR2015 (IC15)~\cite{icdar2015},  
Street View Text Perspective (SVTP)~\cite{SVTP} 
and CUTE80 (CT80)~\cite{CT80}.\end{sloppypar}

\noindent\textbf{Synthetic Datasets} Two public synthetic datasets are employed to train our model: 
\textbf{Synth90K} the $9$-million-word synthetic data released by~\cite{Max2016IJCV} and  \textbf{SynthText} the $8$-million-word data proposed by~\cite{Gupta16}.

\noindent\textbf{IIIT5K}~\cite{MishraBMVC12} is collected from Internet. It has $3000$ cropped word images for test, with nearly horizontal text instances.

\noindent\textbf{SVT}~\cite{Wangkai2011} contains $647$ cropped text images for test. It is collected from Google Street View. Although the text instances are mostly horizontal, many images are severely corrupted by noise and blur, or have very low resolutions. 

\noindent\textbf{IC13}~\cite{icdar2013} has $1095$ regular word patches for test. For fair comparison, we remove images that contain non-alphanumeric characters, which results in $1015$ images. 

\noindent\textbf{IC15}~\cite{icdar2015} consists of images captured incidentally by Google Glasses, and so has many irregular word patches (perspective or oriented). It includes $2077$ images for test. To fairly compare with previous methods~\cite{Cheng2018AON,Luo2019MORAN,ESIR,Cheng2017,cheng_EditDistance,shiPAMI2018}, 
we also used two simplified versions of the IC15 dataset called IC15-Char\&Digit and IC15-$1811$. IC15-Char\&Digit also includes $2077$ images, but discards non-alphanumeric characters in the annotations. IC15-1811 discards the images which have non-alphanumeric characters, and contains $1811$ images.

\noindent\textbf{SVTP}~\cite{SVTP} contains $645$ cropped images for test. Images are selected from side-view angle snapshots in Google Street View, which are mostly perspective distorted.

\noindent\textbf{CT80}~\cite{CT80} consists of $288$ cropped high resolution images for test. It is speciﬁcally collected for evaluating the performance of curved text recognition. 

\subsection{Implementation Details}

The proposed model is implemented using PyTorch. All experiments are conducted on an NVIDIA GTX 1080Ti GPU with $11$GB memory. We use the ADADELTA optimizer~\cite{zeiler2012adadelta} to train the model, with a batch size of $160$. The sampling ratio for Synth90K and SynthText is $1:1$ in one batch and the model is trained $4$ epochs on synthetic datasets. 
The holistic representation dimension is equal to word embedding dimension, which is $512$. So the dimensions $d$ and $d'$ are set to $1024$ and $2048$ respectively in our experiments.

During test phase, borrowed from~\cite{Li_AAAI2019}, we rotate the image $\pm 90$ degrees for images with height twice larger than width. 
The highest-scored recognition result will be chosen as the final output.
Beam search is also applied for the decoder. It keeps the top-$k$ candidates with the highest accumulative scores, where $k$ is empirically set to $5$ in our experiments. 

\subsection{Ablation Study}
\label{sec:ablation}

\paragraph{\bf{CNN Backbone Selection}} \,\,
We first experiment with different CNN models for image encoding, including ResNet18, ResNet34, ResNet50 and ResNet152. Experimental results in Table~\ref{tab:backbone} show that ResNet34 achieves a good balence between model size and accuracy. So we choose ResNet34 as our backbone in the following experiments.

\begin{table}[]
\begin{center}
\scalebox{0.85}{
\begin{tabular}{ c|c|c|c}
\hline
\multirow{2}{*}{Modified CNN Backbone} & \multirow{2}{*}{Image Size} & \multicolumn{2}{c}{Accuracy} \\ \cline{3-4} 
 &  & III5K & IC15 \\ \hline \hline
 ResNet18 & $48\times160$ & $94.0$ & $72.1$ \\ \hline
 ResNet34 & $48\times160$ & $94.7$ & $74.0$ \\ \hline
 ResNet50 & $48\times160$ & $94.8$ & $73.5$ \\ \hline
 ResNet152 & $48\times160$ & $\mathbf{95.0}$ & $\mathbf{74.5}$ \\ \hline
\end{tabular}
}
\end{center}
\caption{Performance with different modified CNN backbones. Modified ResNet34 achieves a good balance between performance and model size.}
\vspace{-1em}
\label{tab:backbone}
\end{table}


\paragraph{\bf{Number of Decoder Blocks}} \,\,
As shown in Row $4$, $6$, $7$ of Table~\ref{tab:Stack_Head_Num}, 
we set the number of decoder blocks to $1$, $2$, $3$ while keeping the number of attention heads as $16$. 
The results show that best performance of our model is achieved when $N = 1$.
This phenomenon is in contrast to the experimental results of the Transformer~\cite{vaswani2017attention}, 
which shows that using more blocks yield better machine translation performance. 

\paragraph{\bf{Number of Attention Heads}} \,\,
Another factor that affects the recognition performance is the number of attention heads $H$.
We evaluate the recognition performance of our models 
with $1$, $4$, $8$, $16$, $32$ attention heads respectively.
The experimental results in Table~\ref{tab:Stack_Head_Num} show that the more attention heads we used, the better performance it can achieved.
In the following, we set the number of attention heads $H$ to $16$.

\begin{table}[]
\begin{center}
\scalebox{0.8}{
{
\begin{tabular}{c|c|c|c}
\hline
\multirow{2}{*}{Block Number ($N$)} & \multirow{2}{*}{Head Number ($H$)} & \multicolumn{2}{c}{Accuracy} \\ \cline{3-4} 
 &  & IIIT5K &  IC15  \\ \hline \hline
$1$ & $1$ & $94.5$ & $72.2$ \\ \hline
$1$ & $4$ & $94.3$ & $73.1$ \\ \hline
$1$ & $8$ & $94.7$ & $\mathbf{74.1}$ \\ \hline
$1$ & $16$ & $\mathbf{94.7}$ & $74.0$ \\ \hline
$1$ & $32$ & $94.5$ & $74.1$ \\ \hline
$2$ & $16$ & $93.7$ & $73.2$ \\ \hline
$3$ & $16$ & $91.6$ & $71.5$ \\ \hline
\end{tabular}}
}
\end{center}
\caption{The performance with different block numbers and attention head numbers in the decoder. It shows that using more heads can slightly improve the performance but using more blocks (with $H = 16$) degrades the performance. 
}
\label{tab:Stack_Head_Num}
\end{table}

\begin{figure*}[]
	\begin{center}
		\includegraphics[width=1.0\textwidth]{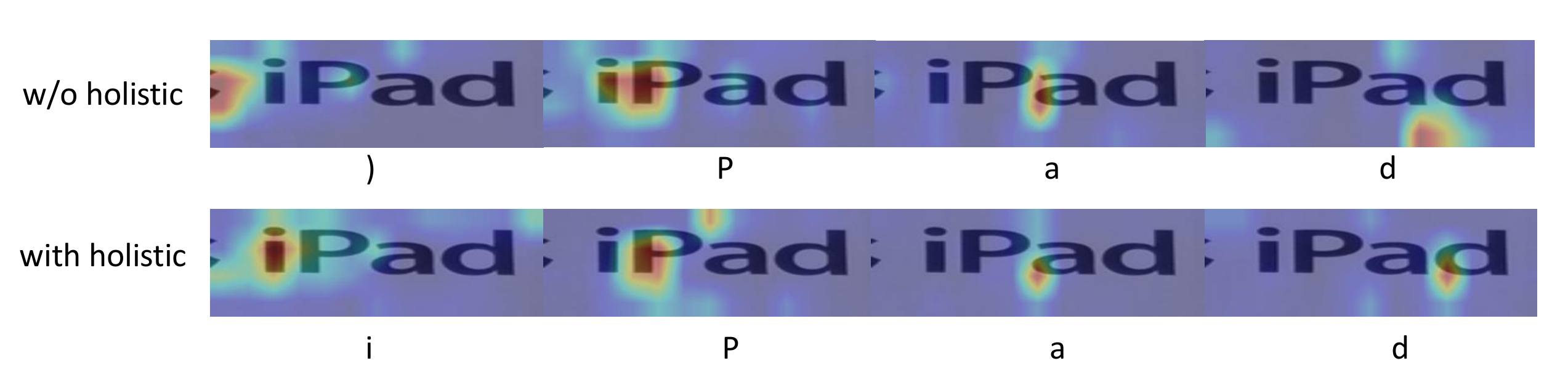}
	\end{center}
	\vspace{-5mm}
	\caption{Case study of attention maps with or without holistic representation. With holistic representation, our algorithm tends to focus more correctly on the regions of characters to be decoded. Best viewed in colour.}
	\label{fig:holisticCase}
\end{figure*}

\paragraph{\bf{Impact of Holistic Representation}} \,\,
The holistic representation vector in our model encodes the rich context information of the entire input image. 
It is fed into the decoder at each time step, together with the last decoded character.
Figure~\ref{fig:holisticCase} demonstrates a case study of the 2D attention maps generated with and without the holistic representation. 
We can see that the algorithm focuses more accurately on the characters to be decoded with the context information provided by the holistic representation.
In addition, we study the effects of the number of bottlenecks that are used to generate the holistic representation. As shown in Table~\ref{tab:Block_Depth}, using more bottleneck modules can improve the performance both for regular and irregular word recognition. When the bottleneck number reachs $6$, the performance is almost saturated. So we set the number of bottlenecks $B$ to $6$ by default.

\begin{table}[]
\begin{center}
\scalebox{0.8}{
{
\begin{tabular}{c|c|c}
\hline
\multirow{2}{*}{Bottleneck Number($B$)} & \multicolumn{2}{c}{Accuracy} \\
\cline{2-3} 
  & IIIT5K &  IC15  \\ \hline \hline
No holistic& $94.2$ & $73.2$ \\ \hline
$2$ & $94.3$ & $73.2$ \\ \hline
$4$ & $94.9$ & $73.8$ \\ \hline
$6$ & $94.7$ & $\mathbf{74.0}$ \\ \hline
$8$ & $\mathbf{94.8}$ & $73.9$ \\ \hline
\end{tabular}}
}
\end{center}
\caption{The performance with different bottleneck number for holistic representation. It shows that using deeper bottleneck can improve the performance both for regular and irregular datasets, and when the bottleneck number is $6$ , the performance is almost saturated.}
\label{tab:Block_Depth}
\vspace{-2mm}
\end{table}

\paragraph{\bf{Impact of Self-Attention}} \,\,
Self-attention plays a key role in many sequence-to-sequence tasks (\emph{e.g.}, machine translation), due to its ability of modeling long-range dependencies.
In the context of image processing, self-attention share a similar spirit with non-local neural networks~\cite{wang2018non}. 
In this section, we examine the impact of self-attention in our proposed model for irregular text recognition. 
We firstly add a self-attention module on top of the convolutional feature maps, to enhance the representation of dependencies between distant image regions.
However, the results in Table~\ref{tab:SlfAttn} show that the addition of self-attention in the encoder does not bring improvement.
On the other hand, to examine the impact of self-attention on the decoder side, we remove the self-attention modules from the decoder.
The recognition performance of the resulting model just moderately drops compared with the original model ($0.7\%$ for IIIT5K containing regular text and $0.1\%$ for IC15 consisting of irregular text), which is still comparable to previous methods.

\begin{table}[]
\begin{center}
\scalebox{0.8}{
{
\begin{tabular}{c|c|c|c}
\hline
Encoder & Decoder & \multicolumn{2}{c}{Accuracy} \\ \cline{3-4} 
Self-attention & Self-attention & IIIT5K & IC15 \\ \hline \hline
$\times$ & $\times$ & $94.0$ & $73.9$ \\ \hline
$\times$ & $\surd$ & $\mathbf{94.7}$ & $\mathbf{74.0}$ \\ \hline
$\surd$ & $\surd$ & $94.5$ & $74.0$ \\ \hline
\end{tabular}}
}
\end{center}
\caption{The performance with or without self-attention in the encoder and decoder. 
Comparing Rows $1$ and $2$, removing self-attention in the decoder from our model results in a moderate performance drop. 
From Rows $2$ and $3$, we can see that adding self-attention in the encoder does not show any improvement.}
\label{tab:SlfAttn}
\vspace{-2mm}
\end{table}

In contrast to machine translation, we find that the usage of self-attention in our irregular text recognizer has 
a relatively small 
impact on the performance.
We analyze that the reasons may be three-fold.
First, the lengths of sequences to be modeled in the irregular text recognition task is typically smaller than that in machine translation.  
For example, in the Multi30K~\cite{elliott-EtAl:2016:VL16} dataset for English-German translation, 
the average lengths of input and output sequences are $11.8$ and $11.1$ respectively.
While in the test set of IC15~\cite{icdar2015}, the average length of output sequences is $5.3$ for irregular text recognition.
Apparently, it is less important to model long-range dependencies for short sequences.
Second, the deep CNN encoder already models a certain level of long-range dependencies, 
given that the receptive field of the final feature layer of ResNet34 is $98$ that is comparable to the input image size ($48 \times 160$).
Last, in machine translation, self-attention is typically used to model the dependencies between words in a sentence or even a paragraph.
There are still rich semantic and syntactic relationships between words that are far from each other.
While for irregular text recognition, each input image usually contains a particular word, 
and the self attention is only used to model character dependencies in a word.
The dependencies between characters of a word are typically weaker than that between words in a sentence or paragraph.
That may be why self-attention does not empirically improve a lot to the performance of irregular text recognition.  

\paragraph{\bf{Impact of Bidirectional Decoder}} \,\,
To evaluate the effectiveness of the bidirectional decoder, we compares the recognition accuracies with different decoders: Normal, which only recognizes text in the left-to-right order; Reversed, which only recognizes text in the right-to-left order; Bidirectional, pick the one with the highest recognition score between the normal and reversed.

\begin{table}[]
\begin{center}
\scalebox{0.8}{
{
\begin{tabular}{c|c|c}
\hline
\multirow{2}{*}{Decoder} & \multicolumn{2}{c}{Accuracy} \\
\cline{2-3} 
  & IIIT5K &  IC15  \\ \hline \hline
Normal & $94.0$ & $73.5$ \\ \hline
Reversed & $94.3$ & $73.3$ \\ \hline
Bidirectional & $\mathbf{94.7}$ & $\mathbf{74.0}$ \\ \hline
\end{tabular}}
}
\end{center}
\caption{Test with different decoders. ``Normal'' means the left-to-right order, ``Reversed'' means right-to-left order and ``Bidirectional'' means the combination of them.}
\label{tab:Bidirectional}
\vspace{-3mm}
\end{table}

Table~\ref{tab:Bidirectional} shows that Normal and Reversed have similar accuracies. Normal outperforms on IC15, while Reversed outperforms on IIIT5K. 
Although there is only little accuracy difference between Normal and Reversed, they have a large performance improvement when combined.

\paragraph{\bf{Impact of Text Rectification}} \,\,
We conducted experiments to study the impact of text rectification over our framework. We used the multi-object rectiﬁcation network in ~\cite{Luo2019MORAN} as the text rectification method.
As Table~\ref{tab:MORN} shows, the text rectification method doesn't work well with our proposed framework. 

\begin{figure}[h]
\small
	\begin{center}
		\includegraphics[width=0.9\linewidth]{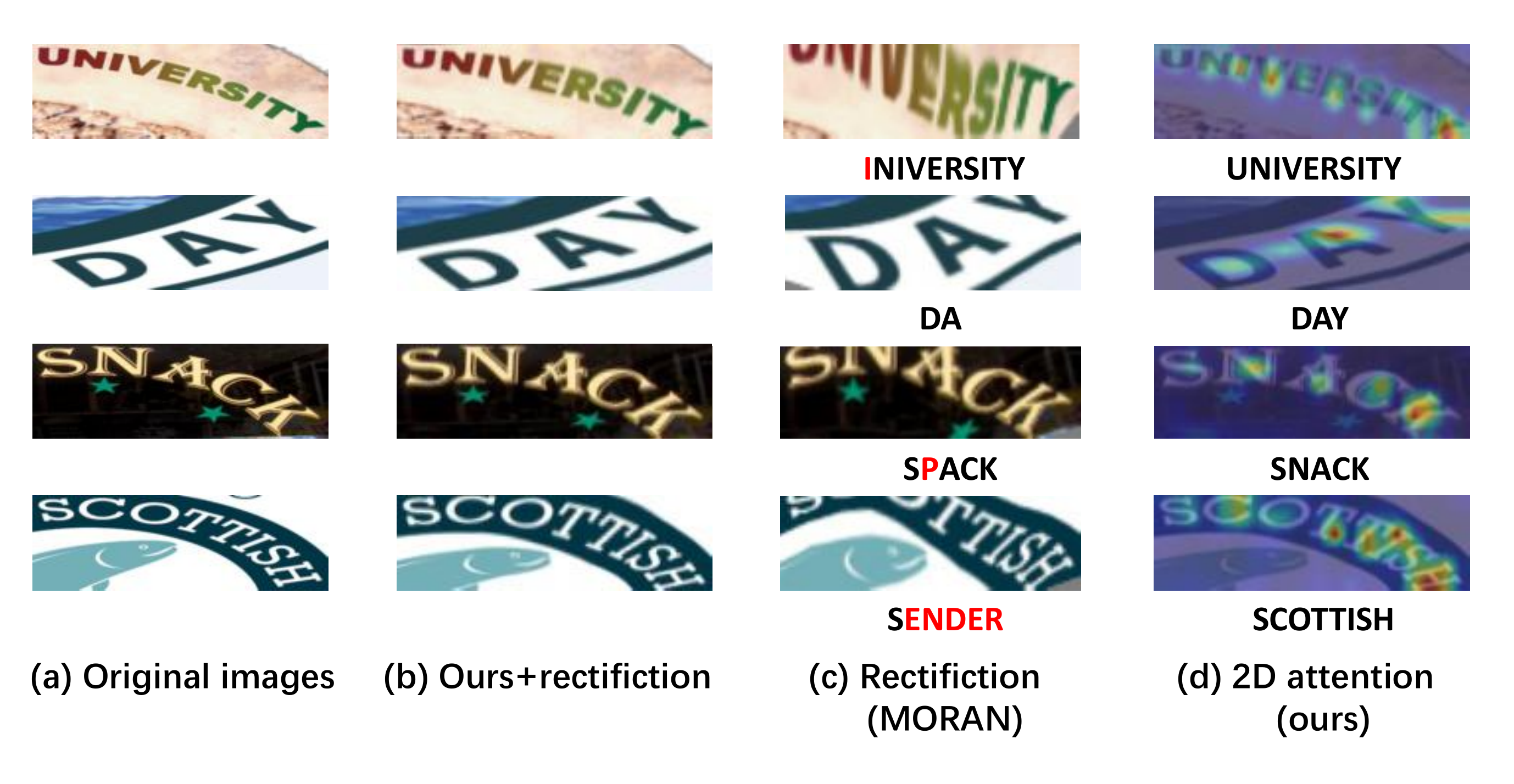}
	\end{center}
    \vspace{-1mm}	
	\caption{ The comparison of our proposed 2D attention based and the rectification based (MORAN~\cite{Luo2019MORAN}) irregular text recognizers. The images in the second column are rectified images by our approach with text rectification method; the third column demonstrates the rectified images and the corresponding predictions using the authors’ implementation; the fourth column gives the predictions of our approach and the heat map by aggregating attention weights at all character decoding steps.
	}
	\vspace{-0.5em}
	\label{fig:rectification}
\end{figure}

The main reason may be that our 2D-attention based approach is already capable of handling irregular texts well. Without rectification, our 2D-attention module can localize individual characters distributed in 2D space. In this manner, rectification has little impact on our approach. As shown in Figure~\ref{fig:rectification}, the text images are significantly rectified by MORAN~\cite{Luo2019MORAN} (a rectification + 1d attention method), but the images are only slightly transformed when our 2d-attention based method is equipped with the same rectification module and trained end-to-end.

\begin{table}[h]
\begin{center}
\scalebox{0.8}{
\begin{tabular}{c|c|c}
\hline
Method    & IIIT5K &  IC15  \\ \hline \hline
Ours      & $\mathbf{94.7}$ & $\mathbf{74.0}$  \\  \hline 
Ours + Rectification & $94.0$ & $74.0$ \\ \hline 
\end{tabular}
}
\end{center}
 \vspace{-1mm}
 \caption{The performance with or without text rectification method on our approach. The text rectification algorithm used the multi-object rectiﬁcation network in ~\cite{Luo2019MORAN}.}
    \label{tab:MORN}
\end{table}

\begin{table*}[!ht]
\begin{center}
\scalebox{0.58}{
\begin{tabular}{r |ccc|cc|c|ccc|ccc|c}
 \hline
\multirow{3}{*}{Method} & \multicolumn{6}{c|}{Regular Text} & \multicolumn{7}{c}{Irregular Text} \\ \cline{2-14} 
 & \multicolumn{3}{c|}{IIIT5K} & \multicolumn{2}{c|}{SVT} & IC13 & \multicolumn{3}{c|}{IC15} & \multicolumn{3}{c|}{SVTP} & CT80 \\ \cline{2-14} 
 & $50$ & $1k$ & None & $50$ & None & None & None & Char\&Digit & 1811 & $50$ & Full & None & None \\
\hline \hline
Yang~\etal 2017~\cite{ijcai2017}*  & $97.8$ & $96.1$ & - & $95.2$ & - & - & - & - & - & $\mathbf{93.0}$ & $\emph{80.2}$ & $75.8$ & $69.3$ \\
Cheng~\etal 2017~\cite{Cheng2017}* & $99.3$ & $97.5$ & $87.4$& $97.1$ & $85.9$ & $93.3$ & - & -& $\mathbf{70.6}$  & - & - & $71.5$ & $63.9$ \\
Liu~\etal 2018~\cite{SqueezeText18}* & $97.0$ & $94.1$ & $87.0$& $95.2$ & - & $92.9$ & - & -& - & $\emph{92.6}$ & $\mathbf{81.6}$ & - & - \\
Liu~\etal 2018~\cite{Liu2018CharNetAC}* &-&-& $92.0$&-& $85.5$ & $91.1$ & $\mathbf{74.2}$ & - & -& - & - & $78.9$ & - \\
Liao~\etal 2019~\cite{Liao2019}* &$\mathbf{99.8}$ &$98.8$ & $91.9$ &$\emph{98.8}$ & $86.4$ & $91.5$ & - & - & - & - & -& - & $79.9$ \\
Yang~\etal 2019~\cite{Yang_2019_ICCV}* & $99.5$&$98.8$ & $\emph{94.4}$& $97.2$ & $88.9$ & $\mathbf{93.9}$ & - & $\emph{78.7}$ & - & - & - & $80.8$ & $\emph{87.5}$ \\
Liao~\etal 2019~\cite{Minghui2019Mask}* & $\mathbf{99.8}$&$\mathbf{99.3}$ & $\mathbf{95.3}$& $\mathbf{99.1}$ & $\mathbf{91.8}$ & $\mathbf{95.0}$ & - & $78.2$ & - & - & - & $\emph{83.6}$ & $\mathbf{88.5}$ \\
Wan~\etal 2020~\cite{wan2020textscanner}* & $99.7$&$\emph{99.1}$ & $93.9$& $98.5$ & $\emph{90.1}$ & $92.9$ & - & $\mathbf{79.4}$ & - & - & - & $\mathbf{84.3}$ & $83.3$ \\
\hline

Lee and Osindero 2016~\cite{Lee_2016_CVPR}$^{\,\,\,}$ & $96.8$& $94.4$& $78.4$& $96.3$ & $80.7$ & $90.0$ & - & - & - & - & - & - & - \\
Wang and Hu 2017~\cite{OCRNIPS17}$^{\,\,\,}$ &$98.0$ &$95.6$ & $80.8$ &$96.3$ & $81.5$ & - & - & - & - & - & - & - & - \\
Shi~\etal 2016~\cite{shiCVPR2016}$^{\,\,\,}$ &$96.2$ &$93.8$ & $81.9$ &$95.5$ & $81.9$ & $88.6$ & - & -& -  & $91.2$ & $77.4$ & $71.8$ & $59.2$ \\
Liu~\etal 2016~\cite{BMVC2016_43}$^{\,\,\,}$ &$97.7$ &$94.5$ & $83.3$& $95.5$& $83.6$ & $89.1$ & - & -& - & $94.3$ & $83.6$ & $73.5$ & - \\
Shi~\etal 2017~\cite{ShiBY15}$^{\,\,\,}$ &$97.8$ &$95.0$ & $81.2$ & $\emph{97.5}$& $82.7$ & $89.6$ & - & - & -& $92.6$ & $72.6$ & $66.8$ & $54.9$ \\
Bai~\etal 2018~\cite{cheng_EditDistance}$^{\,\,\,}$ &$99.5$ &$97.9$ & $88.3$ &$96.6$ & $87.5$ & $\mathbf{94.4}$ & -& - & $73.9$ & - & - & - & - \\
Cheng~\etal 2018~\cite{Cheng2018AON}$^{\,\,\,}$ &$\emph{99.6}$ &$98.1$ & $87.0$ &$96.0$ & $82.8$ & - & - & $68.2$ & - & - & - & - & $76.8$ \\
Shi~\etal 2018~\cite{shiPAMI2018}$^{\,\,\,}$  & $\emph{99.6}$& $\emph{98.8}$& $93.4$& $97.4$& $\emph{89.5}$ & $91.8$ & - & -& $\emph{76.1}$ & $94.0$ & $83.7$ & $78.5$ & $79.5$ \\
Gao~\etal 2019~\cite{gao2019reading}$^{\,\,\,}$ & $99.1$& $97.9$& $81.8$  & $97.4$& $82.7$ & $88.0$ & - & - & - & -& - & - & - \\
Li~\etal 2019~\cite{Li_AAAI2019}$^{\,\,\,}$ & -& -& $91.5$& -& $84.5$ & $91.0$ & $\emph{69.2}$ & - & - & -& - & $76.4$ & $\emph{83.3}$ \\
Luo~\etal 2019~\cite{Luo2019MORAN}$^{\,\,\,}$ &$97.9$ &$96.2$ & $91.2$ &$96.6$ & $88.3$ & $92.4$ & - & $68.8$ & - & $\emph{94.3}$ & $\emph{86.7}$ & $76.1$ & $77.4$ \\
Zhan~\etal 2019~\cite{ESIR}$^{\,\,\,}$ & $\emph{99.6}$&$\emph{98.8}$ & $93.3$ &$97.4$ & $\mathbf{90.2}$ & $91.3$ & - & $\emph{76.9}$ & - & -& - & $79.6$ & $83.3$ \\
Wang~\etal 2020~\cite{wang2020decoupled}$^{\,\,\,}$ & $-$&$-$ & $\emph{94.3}$ &- & $89.2$ & $\emph{93.9}$ & - & $74.5$ & - & -& - & $\emph{80.0}$ & $\emph{84.4}$ \\

Ours$^{\,\,\,}$ &$\mathbf{99.8}$ &$\mathbf{99.1}$ & $\mathbf{94.7}$ & $\mathbf{97.7}$ & $88.9$ & $93.2$ & $\mathbf{74.0}$ & $\mathbf{77.1}$ & $\mathbf{79.5}$ & $\mathbf{95.2}$ & $\mathbf{89.5}$ & $\mathbf{80.9}$ & $\mathbf{85.4}$ \\
\hline
\end{tabular}
}
 \end{center}
 \vspace{-1mm}
 \caption{Scene text recognition performance on public datasets. ``Char\&Digit" means discard non-alphanumeric characters in the prediction and annotation, ``$1811$" means discard the images which have any non-alphanumeric characters, and there are $1811$ images left. ``$50$", ``$1k$" and ``Full" are lexicon sizes, ``None" means no lexicon. For datasets with lexicons, we select from lexicon the one with the minimum edit distance to the predicted word. ``*" indicates models trained with both word-level and character-level annotations. \textbf{Bold} and \emph{Italic} fonts represent the best and second best performance respectively. }
    \label{tab:Comparison}
\end{table*}

\subsection{Comparison with State-of-the-art}


In this section we evaluate our model with $N = 1$, $H = 16$ and $d = 1024$, in comparison with state-of-the-art approaches on several benchmarks. For fair comparison, we only demonstrate the performance of SAR~\cite{Li_AAAI2019} and TextScanner~\cite{wan2020textscanner} trained with synthetic data. 
As shown in Table~\ref{tab:Comparison}, 
our proposed method outperforms other word-level approaches on \textbf{all of} evaluated settings for irregular text recognition. In particular, it achieves accuracy increases of $3.4\%$ (from $76.1\%$ to $79.5\%$) on IC15-1811 and $1.0\%$ (from $84.4\%$ to $85.4\%$) on CT80.

And for regular text datasets, our performance is also competitive. 
On the IIIT5K dataset which contains the largest number of test images over the three evaluated regular datasets, our model is $0.4\%$ better than the best word-level model ($94.7\%$ v.s. $94.3\%$).
Although our method only uses word-level annotation, it shows a little worse than the best character-level model on IIIT5K (($94.7\%$ v.s. $95.3\%$)).

We also compare the model size and computation speed of our model with a simple yet strong baseline~\cite{Li_AAAI2019} and a state-of-the-art model~\cite{shiPAMI2018}. The experiment is performed on a $1080$Ti GPU with a batch size of $20$.
Due to the non-recurrence property, our model is significantly faster than these two RNN-based models. 

\begin{table}[]
\begin{center}
\scalebox{0.75}{
\setlength{\tabcolsep}{1mm}{
\begin{tabular}{c|c|c|c}
 \hline
Method & Model Size & \begin{tabular}[c]{@{}c@{}}Forward Time\\ per Batch\end{tabular} & \begin{tabular}[c]{@{}c@{}}Backward Time\\ per Batch\end{tabular} \\
 \hline \hline
Shi~\etal 2018~\cite{shiPAMI2018} & $22$M &  $65$ms & $143$ms \\ \hline
Li~\etal 2019~\cite{Li_AAAI2019} & $61$M &  $404$ms & $903$ms \\ \hline
Ours & $80$M & $\mathbf{51}$ms & $\mathbf{96}$ms \\ \hline
\end{tabular}}
}
\end{center}
\caption{The comparison on training speed and model size. The speed is evaluated with $20$-sized batches in average. 
Our model is $1.5\times$ to $9.4\times$ faster in backward pass and $1.3\times$ to $7.9\times$ faster in forward pass. }
\vspace{-2em}
 \label{tab:Speed}
\end{table}

\begin{figure}[]
\small
	\begin{center}
		\includegraphics[width=8.1cm, height=6.5cm]{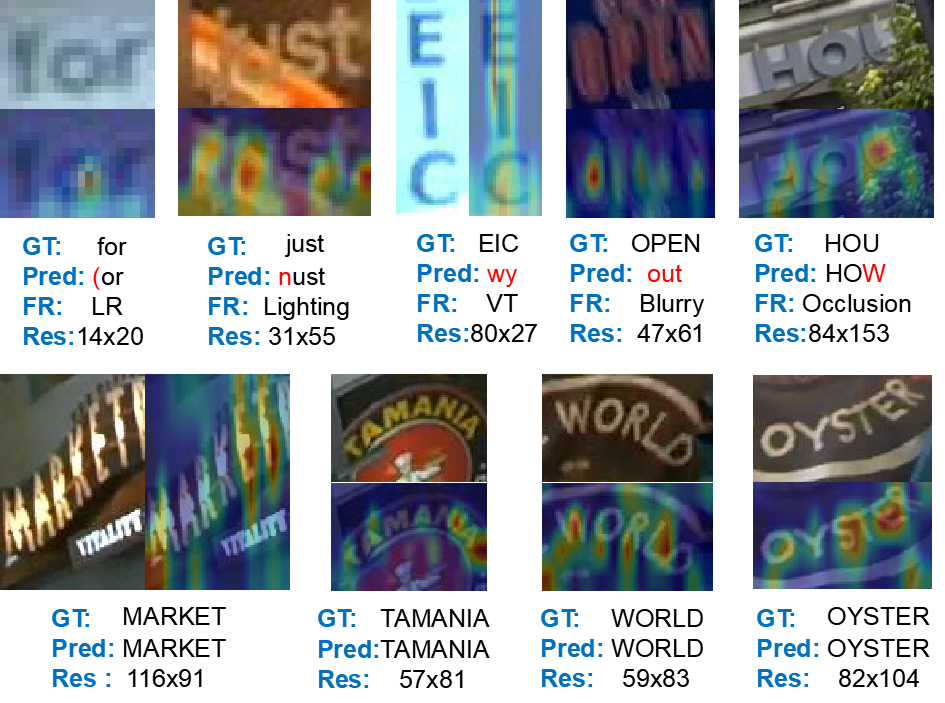}
	\end{center}
    \vspace{-1mm}	
	\caption{Some success and failure cases by our approach. The $2$D attention weights combining all decoding steps are also illustrated. ``GT": Ground Truth, ``Pred": Prediction, ``FR": Failure Reason, ``Res": Original Image Resolution. The reasons for failure include blurry, low resolution (LR), lighting, vertical text (VT), and occlusion \etc. Best viewed in colour.}
	\vspace{-0.5em}
	\label{fig:FailureCase}
\end{figure}

Some success and failure cases are also presented in Figure~\ref{fig:FailureCase}. In the second line, those pictures with different shapes can be correctly recognized, it shows that our model is capable of dealing with text of complex shapes.
There are several reasons for our method to make wrong decisions, including blurry images, low resolution, vertical text, lighting and occlusion. We found that the recognition accuracy for vertical text is very low, the main reason maybe that there is few vertical text samples in current synthetic datasets.

\begin{figure}
\begin{minipage}[h]{1\linewidth}
\centering
\includegraphics[width=1\linewidth]{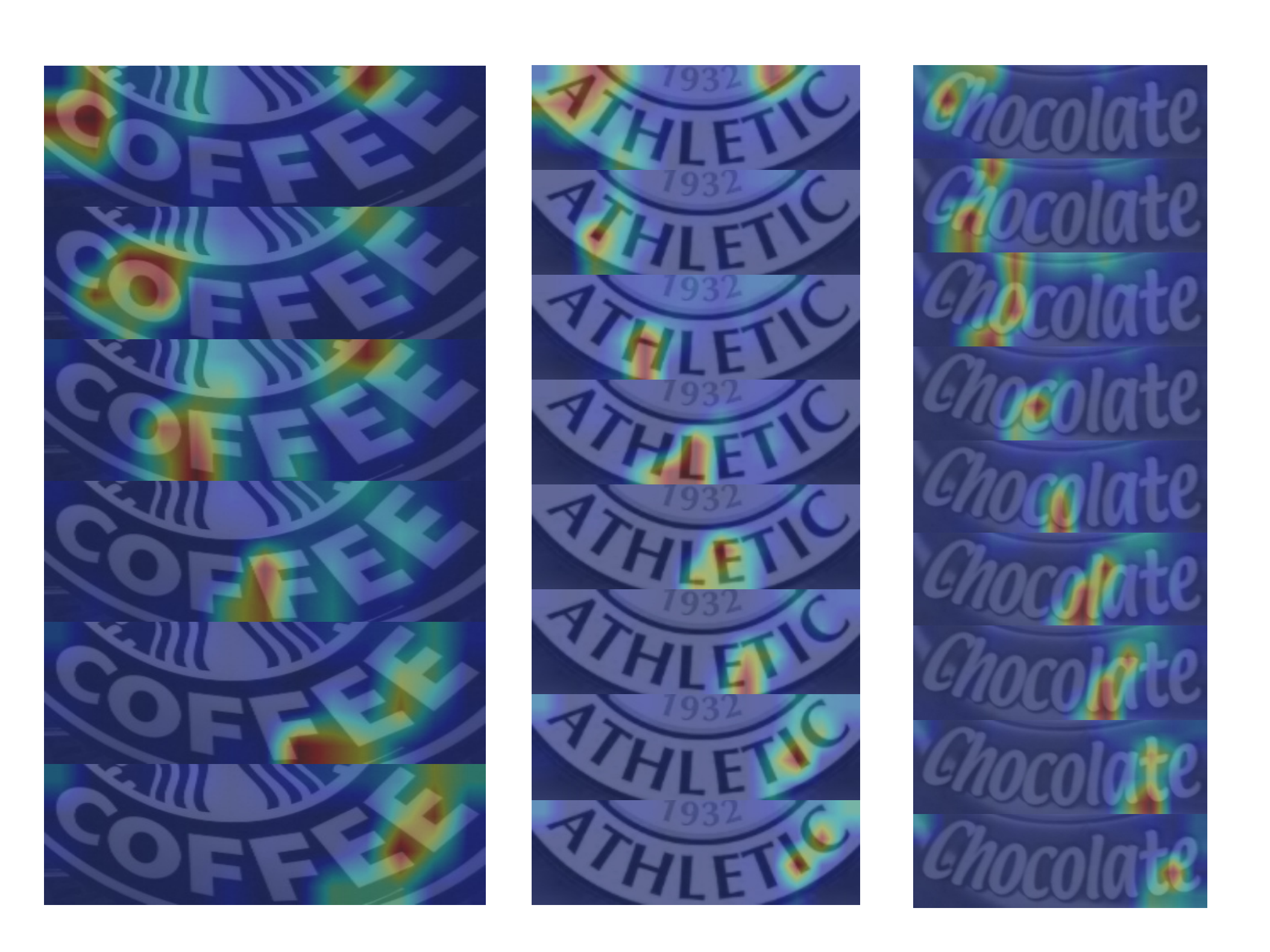}
\end{minipage}

\begin{minipage}[h]{1\linewidth}
\centering
\includegraphics[width=1\linewidth]{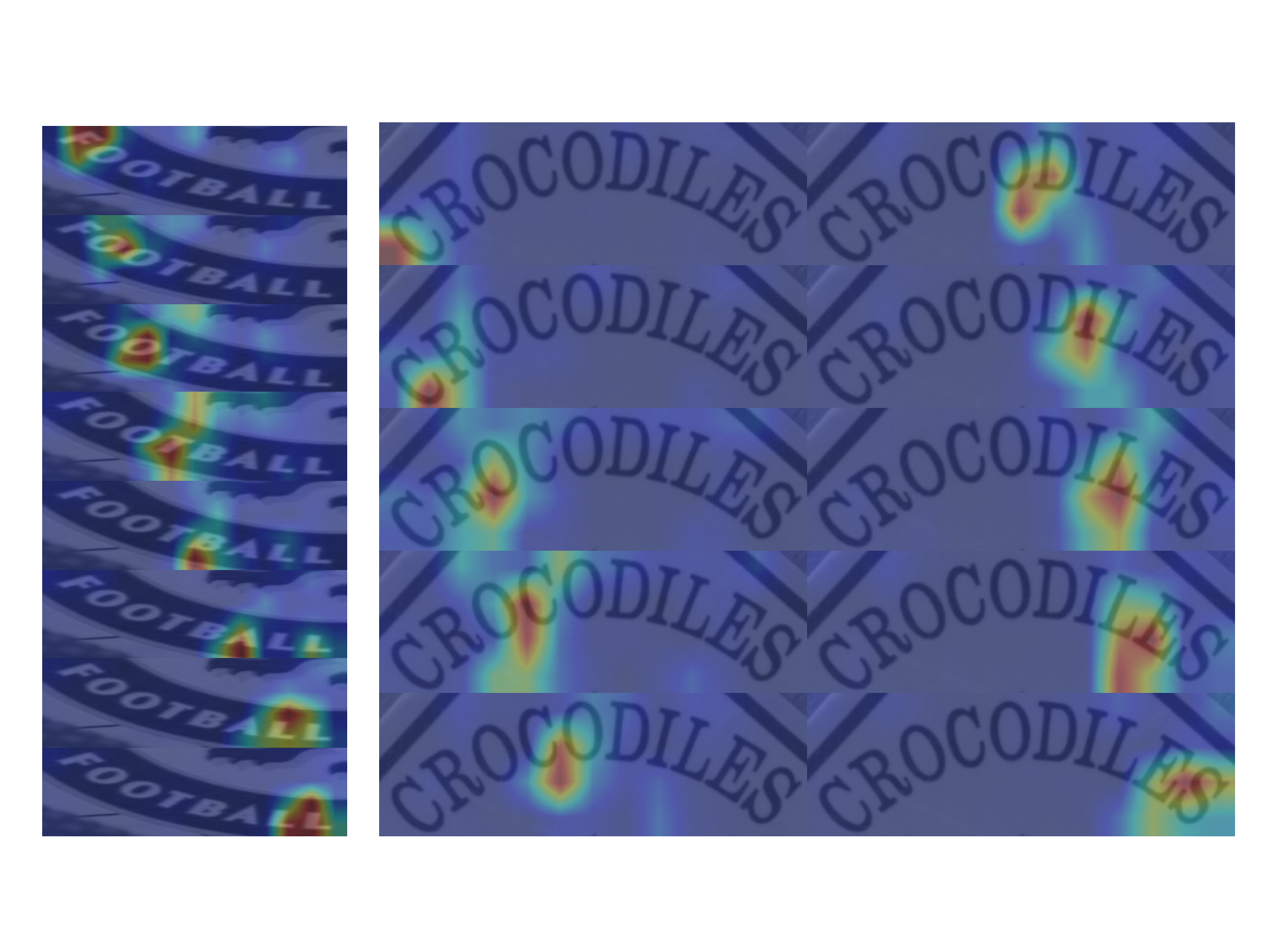}
\end{minipage}

\caption{ Case study of the attention map at different decoding steps}
\label{fig:steps}
\end{figure}

Figure~\ref{fig:steps} shows some cases of the attention map at different decoding steps. We can see that the decoder can focus on the right region to the characters to be decoded, but sometimes there are little position drift.

\section{Conclusion}

In this work, we propose a simple and strong holistic representation guided attention network for scene text recognition.
The simplicity of our model is reflected in  three aspects. 
1) Simple architecture: the proposed model directly connects a CNN encoder to an attention-based encoder.
We do not convert input images into sequences as in many existing irregular text recognizers.  
2) Parallel training: as a non-recurrent network, our model can be trained in parallel.
Compared with two state-of-the-art RNN-based irregular text recognizers, the computational speed of our model is significantly faster. 
3) Simple training data: our model only relies on the word-level annotations.
And we use the holistic representation of the image in the decoder to help the $2$D attention focus on the correct region.
As a simple meta-algorithm, this model can be extended in multiple ways, such as incorporating multi-scale image features via stacked $2$D attention and resizing input images while keeping aspect ratios. We leave them for further work.

\bibliography{mybibfile}

\end{document}